\documentclass[10pt,twocolumn]{article} 
\usepackage{simpleConference}
\usepackage{times}
\usepackage{graphicx}
\usepackage{amssymb}
\usepackage{url,hyperref}

\usepackage{booktabs} % For formal tables
\usepackage{url}
\usepackage{graphicx}
\usepackage{algorithm}
\usepackage{algpseudocode}
\usepackage{subcaption}
\usepackage{bm}
\usepackage{balance}
\usepackage{amsmath}
\usepackage[
backend=biber,
style=numeric,
sorting=ynt
]{biblatex}

\begin{document}

\title{Autostacker: A Compositional Evolutionary Learning System}

\author{Boyuan Chen \\
Columbia University \\
bchen@cs.columbia.edu \\
\and
Harvey Wu \\
Columbia University \\
wu.harvey@columbia.edu
\and
Warren Mo \\
University of Chicago \\
warrenmo@uchicago.edu
\and
Ishanu Chattopadhyay \\
University of Chicago \\
ishanu@uchicago.edu
\and
Hod Lipson \\
Columbia University \\
hod.lipson@columbia.edu
}

\maketitle
\thispagestyle{empty}

\begin{abstract}
We introduce an automatic machine learning (AutoML) modeling architecture called Autostacker, which combines an innovative hierarchical stacking architecture and an Evolutionary Algorithm (EA) to perform efficient parameter search. Neither prior domain knowledge about the data nor feature preprocessing is needed. Using EA, Autostacker quickly evolves candidate pipelines with high predictive accuracy. These pipelines can be used as is or as a starting point for human experts to build on. Autostacker finds innovative combinations and structures of machine learning models, rather than selecting a single model and optimizing its hyperparameters. Compared with other AutoML systems on fifteen datasets, Autostacker achieves state-of-art or competitive performance both in terms of test accuracy and time cost.
\end{abstract}

\section{Introduction}
Wolpert's No Free Lunch theorem \cite{wolpert1992stacked} implies that no model can be expected to generalize well to all data. Machine Learning practitioners, upon encountering each new dataset, must ask:  \emph{What models can we use, and how can we pick the best hyperparameters for our chosen model?}
A successful choice of model often requires considerable experience and knowledge; good choices of hyperparameters often come as the product of time-intensive tuning. 
Automating both parts of the modeling procedure - model selection and hyperparameter optimization - would make the fruits of machine learning accessible to a wider community, making it highly desired in both academia and industry.  \\\\
An AutoML system aims to do just that: providing an automatically generated baseline to make it easier to solve machine learning problems.
Such a system takes in a formatted dataset as input and outputs one or more modeling pipelines that achieve reasonable performance on the dataset.
Recent efforts in AutoML, such as AutoSklearn \cite{feurer2015efficient} and TPOT \cite{OlsonGECCO2016} demonstrate success in a variety of datasets.

\begin{figure}[t]
  \centering      
  \includegraphics[width=1.0\linewidth]{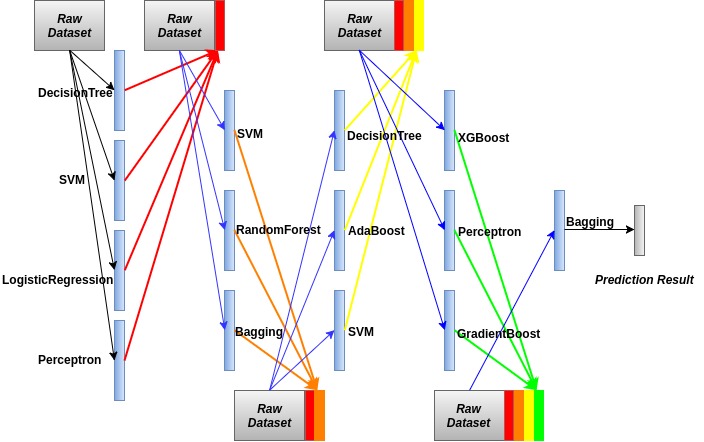}
  \caption{A typical pipeline generated by Autostacker. Each column represents a layer. Each node in a layer represents a machine learning primitive model (e.g, SVM, MLP). The number of layers and nodes per layer can be specified beforehand or treated as a hyperparameter. The raw dataset is used as input for the first layer. In the following layers, the prediction results from each node will be added to the raw dataset as synthetic features (new colors). The new dataset generated by each layer is fed as input to the next layer.}
  \label{fig:system architecture}
\end{figure}

\begin{figure*}[t]
\vspace{0mm}
\centering
    %\captionsetup[subfigure]{aboveskip=-8pt} % spaceing between image and caption for top row
	\begin{subfigure}[t]{0.9\linewidth}
		\centering
		\includegraphics[width=\linewidth,keepaspectratio]{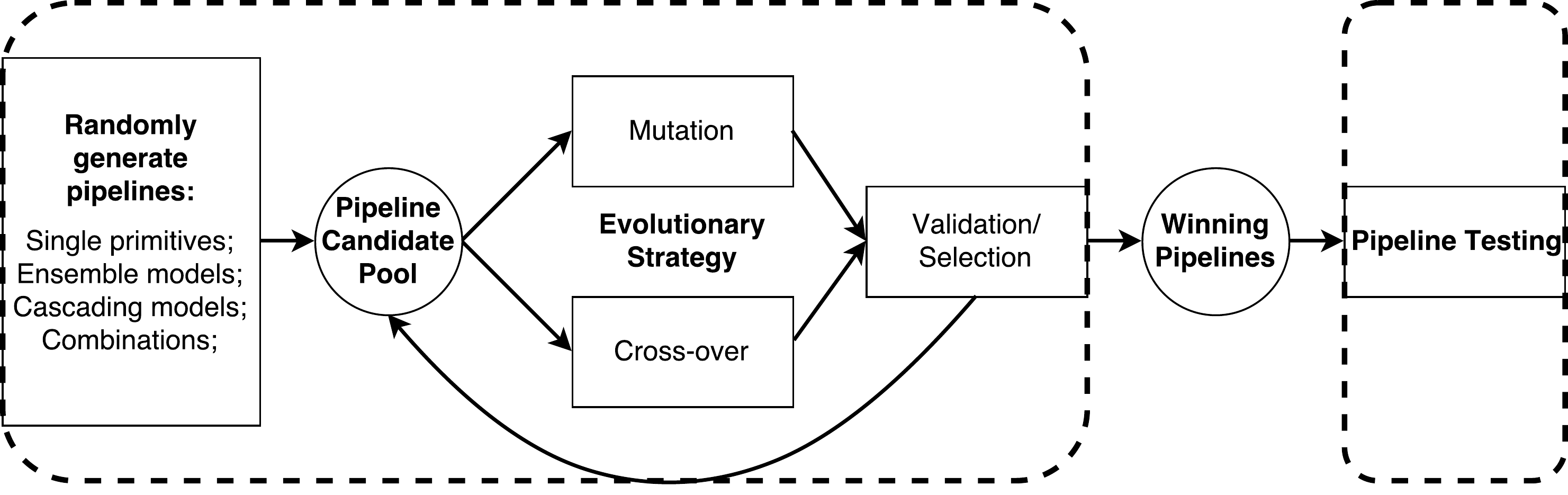}
		\label{fig:process_1}
	\end{subfigure}

\caption{An overview of pipeline generation. We randomly generate initial pipelines and feed those into the basic EA algorithm, looping the process to generate winning pipelines. The hyperparameters of each pipeline (number of layers and nodes) can be explicitly defined by the user or tuned by Autostacker.} 
\label{fig:process} 
\end{figure*}

In this work, we present an AutoML architecture called Autostacker. 
Inspired by the stacking method \cite{wolpert1992stacked}\cite{breiman1996stacked} of ensemble learning, Autostacker automatically discovers pipelines made up of one or many models. 
Compared to other AutoML frameworks, Autostacker demonstrates competitive performance in both accuracy and time when evaluated on fifteen datasets.
The following three properties of Autostacker allow it to generalize well to new data:  

\begin{itemize}
\item \textbf{Cascading} Despite the rise of "big data", many datasets are still small and sparse. We tackle this challenge through cascading: always using the original dataset in all stacking layers while concatenating synthetic features in each stacking layer. More details are provided in the Approach section below. \\

\item \textbf{Model Flexibility} Existing AutoML frameworks generate a full pipeline that includes data preprocessing, feature engineering, and model selection. Model selection usually involves the optimization of a single machine learning primitive, such as a Support Vector Machine (SVM) \cite{cortes1995support}, or a traditional ensemble method, such as Boosting \cite{kearns1988learning}\cite{schapire1990strength}\cite{freund1995boosting}. Autostacker allows for flexible combinations of many machine learning primitives, resulting in a larger search space. \\

\item \textbf{Evolutionary Search Algorithm} EAs allow us to tractably find good solutions in a large space of variables \cite{eiben}. Such variables include the type of primitive machine learning models, the configuration settings of the framework (for instance, the number of primitive models in each stacking layer) and the hyperparameters in each primitive model. In our work, we consider all of the elements above as hyperparameters. Instead of treating AutoML as an optimization problem \cite{feurer2015efficient}, we model it as a search problem in this large space of hyperparameters. Exploiting the parallel nature of Evolutionary Algorithms, Autostacker quickly finds good candidate pipelines. As shown in the Results section, we achieve competitive performance with only a very basic version EA. 

\end{itemize}

\section{Related Work}

\subsection{Stacking and Cascading}
Stacking is a decades-old method of ensemble learning \cite{wolpert1992stacked}. The first layer takes in the original dataset; the next layer is fed the outputs of the classifiers in the first layer, and so on so forth. Intuitively, the later layers can learn the mistakes that classifiers in the previous layers make, and correct them. The related approach of cascading - taking the output of one model and feeding it into another - was first explored as an ensemble learning technique in the work of Viola and Jones \cite{viola2001rapid}. Data is fed through a series of binary classifiers. If a classifier outputs true, the data travels to the next classifier; if a classifier outputs false, the iteration ends and the cascade returns false. If the last classifier is reached and outputs true, the cascade returns true. Cascaded Classification Models (CCMs), a more sophisticated approach to cascading, was introduced by Heitz et al. \cite{heitz2009cascaded} as a way to decompose the complex problem of scene understanding into component problems. We believe that neither stacking nor cascading have been explored in the AutoML literature.

\subsection{Automatic Machine Learning}
AutoML research has focused on combining two tasks: machine learning pipeline building and intelligent model hyperparameter search. Auto-Weka \cite{thornton2013auto}\cite{kotthoff2016auto} selects a single machine learning primitive and optimizes its hyperparameters. Auto-Weka is built on top of Weka \cite{hall2009weka}, and uses Bayesian Optimization (Sequential model-based optimization) to search for optimal hyperparameter settings of the pipeline. The pipeline here follows the traditional machine learning work process: from data preprocessing, feature engineering to single model prediction. However, fixed order pipelines, especially with a single model prediction, are not suitable for complicated problems or small sample datasets. AutoSklearn \cite{feurer2015efficient} follows a similar methodology as above, using the scikit-learn \cite{pedregosa2011scikit} machine learning library as a toolbox, as well as Bayesian Optimization to tune hyperparameters. 

There are also several works on Bayesian Optimization which are designed specifically for large scale parameter configuration problems like AutoML. For example, RoBO  \cite{springenberg2016bayesian} includes multiple implementations of different Bayesian Optimization algorithms with the flexibility of changing the components of this process. Hyperopt \cite{bergstra2013hyperopt} takes advantage of Sequential model-based optimization and considers the choice of classification models and preprocessing models together as an integral optimization problem. Other Bayesian approaches for large scale parameter search include SMAC \cite{hutter2011sequential} and Spearmint \cite{snoek2012practical}.

The use of EAs to perform hyperparameter optimization in an AutoML setting was recently explored in the TPOT architecture \cite{olson2016automating}. Extending the traditional "data scientists'" pipeline used in AutoWeka and Autosklearn, TPOT allows for parallel feature engineering prior to model prediction. Subsequently, TPOT uses Evolutionary Algorithms to treat the parameter configuration problem as a search problem.

All of the aforementioned approaches, however, focus on configuring a single machine learning primitive, with traditional ensemble architectures as a supplement. AutoSklearn allows ensemble models to be built on the fly but it only considers traditional ensemble approaches. Ensemble Selection \cite{caruana2004ensemble} was found to have robust and efficient performance while stacking \cite{wolpert1992stacked} and gradient-free numerical optimization tended to be less efficient and to easily overfit \cite{feurer2015efficient}.

Autostacker, on the other hand, is an ensemble method by default. It handles single model and ensemble approaches simultaneously as basic primitives. The cascading architectures generated by Autostacker allow 
synergistic combinations of ML primitives to "correct each others mistakes" and improve generalization. Moreover, Autostacker allows multiple ensemble models to be used in the same architecture. We believe that Ensemble Learning deserves deeper consideration in the AutoML process, as it is generally more robust and can outperform individual models most of the time \cite{opitz1999popular}\cite{polikar2006ensemble}\cite{rokach2010ensemble}. 

Thus, instead of taking the traditional route of designing an AutoML system that learns to choose a single model and optimize it, we encourage Autostacker to find innovative combinations or arrangements of ML primitives. We hypothesize that this model flexibility is a major factor in Autostacker's empirical success when compared to other AutoML systems. 

However, by stacking models on top of each other, our search space is much larger than that of single-model AutoML systems such as TPOT or AutoSklearn. Naturally the primitives in a candidate pipeline need to be optimized as well, further compounding our problem. We tackle this issue by using a basic Evolutionary Algorithm, rather than Bayesian Optimization, to find suitable hyperparameters. EAs have recently seen a renaissance in other fields of machine learning such as neural network optimization and reinforcement learning \cite{Morse}\cite{such}, confirming their status as an optimization workhorse when dealing with large search spaces. We note that TPOT also uses EAs to perform parameter search.

\section{METHODS}

\subsection{Problem Setting}

In supervised learning, a model acts as a mapping function $f^{i}$ from the sample input data $\textbf{X}$ to the output data $\textbf{Y}$: 

\begin{equation}
    \textbf{Y} \leftarrow f_{\textbf{H}, \boldsymbol{\Theta}}^{i}(\textbf{X})   
\end{equation}
The model $f^i$, belonging to a family of models $\boldsymbol{F}$, is governed by two parameters. Here we call $\textbf{H}$ hyperparameters, and $\boldsymbol{\Theta}$ model parameters. In AutoML, the focus lies in choosing the appropriate $f^i$ and finding good $\textbf{H}$; $\boldsymbol{\Theta}$ is delegated to the training process. The scope of $\textbf{H}$ varies between different systems.

Furthermore, to make the definition of the problem clear, we will use the terminology listed below throughout this paper:

\begin{itemize}

\item Primitive and Pipeline: primitive denotes an existing machine learning model, such as a DecisionTree.
In addition, these also include traditional ensemble learning models, such as AdaBoost and Bagging.
The pipeline is the output of Autostacker, which is a single primitive or a combination of primitives.

\item Layer and Node: Figure \ref{fig:system architecture} shows the architecture of Autostacker, which is formed by multiple stacking layers and multiple nodes in each layers.
Each node represents a machine learning primitive model.

\end{itemize}

\subsection{System Architecture}

The working process of Autostacker is shown in Figure \ref{fig:process} and a sample pipeline built by Autostacker is shown in Figure \ref{fig:system architecture}. Each pipeline consists of multiple layers, where each layer contains multiple nodes. These nodes are primitive machine learning models. The $i$th layer takes in the dataset $X_i$, and outputs the prediction result $Y_{i, j}$, where $Y_{i, j}$ denotes the prediction result of the $j$th node in the $i$th layer ($i = 0, 1, 2, ..., I$, $j$ = $0, 1, 2, ..., J$). After each layer's prediction, we add these prediction results back to the dataset used as input to the layer as synthetic features, and then use this newly generated dataset as the input of the next layer. In other words, the input of $ith$ layer $X_i$ is updated as following:

\begin{equation}
    X_i = X_{i-1} \cup Y_{i-1, 0} \cup Y_{i-1, 1} \ldots \cup Y_{i-1, J'}
\end{equation}
where $J'$ is the number of nodes in $(i-1)th$ layer. With each new layer, the dataset gets more and more synthetic features until the last layer which only consists of a single node. We take the output of the last layer as the final output of this machine learning problem.

Again, if we use $f_k$ to denote the $k$th ($k = 0, 1, 2, ..., K$) feature in the dataset, the final dataset will contain
\begin{equation}
    (K + 1) + \sum_{i=0}^{I-1}(N_{i}+1)
\end{equation}
features in total and this new dataset will be used in the last layer prediction. $N_{i}$ (0,1,2,...) is the number of nodes in the $i$th layer. The total number of features in the dataset before the last layer can be specified by users.

Unlike the traditional stacking algorithm in ensemble learning, which only feeds the prediction results into next layer as inputs, this proposed architecture always cascades the information directly from the raw dataset. To our best knowledge, it is the first time that this algorithm is generalized and incorporated into AutoML system, although similar methods have been tried in practice to solve specific machine learning problems. Here are the considerations:

\begin{itemize}

\item The number of items in the dataset could be very small. If so, the prediction result from each layer could contain very little information about the problem and it is very likely that the primitives bias the outcomes a lot. Accordingly, throwing away the raw dataset could lead to high-biased prediction results which is not suitable for generalization, especially for situations where we could have more training data in the future.

\item Moreover, by combining the new synthetic features with the raw dataset, we implicitly give some features more weight when these features are important for prediction accuracy. Yet we do not throw away the raw dataset, as in regular stacking, because we do not fully trust the primitives in each individual layers. We can consequently reduce the influences of bias coming from individual primitive and noise coming from the raw dataset. 

\end{itemize}

The hyperparameter space in Autostacker consists of four parts:
\begin{equation}
    \boldsymbol{H} = \begin{cases}
      \text{type of each primitive}\\
      \text{each model hyperparameter within each primitive}\\
      \text{number of layers in each pipeline}\\
      \text{number of nodes in each layer}
    \end{cases}
\label{hyperparameterspace} 
\end{equation}

The following attributes of Autostacker can be configured by the user based on their computational resources and/or time constraints:
\begin{itemize}
\item $I$ and $J$: the maximum number of layers and the maximum number of nodes corresponding to each layer.

\item The types of the primitives. Here we provide a dictionary of primitives which only serves as a search space. Additional primitives can be added.

\end{itemize}

Note that Autostacker provides two ways of specifying $I$ and $J$. The default mode is to let users simply specify the maximum range of $I$ and $J$. Only two positive integers are needed to enable Autostacker to explore different configurations. There are two advantages here: 1. This mode frees the system of constraints and allows for the discover of further possible innovative pipelines. 2. The search process achieves a significant speedup. We will illustrate this point in the Experiment section later.

Another choice is to explicitly denote the value of $I$ and $J$. This allows systems to build pipelines with a specific number of layers and number of nodes per layer based on allowed computational power and time.

The search algorithm for finding the appropriate hyperparameters is described in the next section.

\subsection{Search Algorithm}
In this paper, a basic Evolutionary Algorithm (EA) has been chosen as the search algorithm to find the group of hyperparameters $\boldsymbol{H}$ which create better model pipelines. Our bare-bones EA only involves mutation and cross-over, with no sophisticated techniques. As we will show later, our system can already achieve significantly better performance with this straightforward baseline algorithm. Algorithm \ref{phc} provides the details of this algorithm in our system.

First, we generate $N$ completed pipelines by randomly selecting the hyperparameters. Then we run a one-step mutation on top of half of these pipelines to get another $N / 2$ pipelines. The candidates for mutation are chosen randomly. We then use another $N / 2$ pipelines to run cross-over. By now, we can already get another new $N$ pipelines in total.

The one-step mutation randomly changes one of the hyperparameters in $\boldsymbol{H}$ as in set (\ref{hyperparameterspace}). One example could be the number of estimators in a Random Forest Classifier, or replacing a SVM classifier with a logistic regression classifier. Cross-over exchange part of a pair of pipelines' topology. For example, we can take the first half of the layers in one pipeline and the second half of the layers in another pipeline to formalize a new pipeline.

Now we train these $2N$ pipelines and evaluate them through cross validation. Then $N$ pipelines with the highest validation accuracies are selected as the seed pipelines for the next generation of mutation and cross-over. Once the seed pipelines are ready, another one-step mutation and cross-over will be applied on them and another round of evaluation and selection will be executed afterwards. The same loop continues until the end of all the iterations, where the number of iterations $M$ can be specified by users.

% Insert the algorithm
\begin{algorithm}
\caption{Basic EA Search}
\label{phc}
\begin{algorithmic}[1]
\State $N = 200$
\State $M = 10$
\State $iter\_init = Random(N)$
\For {iter in $M$}
    \State Randomly sperate $iter\_init$ into two equal parts, we get
    \State $iter\_init\_1$ and $iter\_init\_2$.
    \State $new\_gen\_1 = MUTATION (iter\_init\_1)$
    \State $new\_gen\_2 = CROSSOVER (iter\_init\_2)$
    \State $new\_gen = new\_gen\_1 + new\_gen\_2$
    \State $eva\_pip = iter\_init \cup new\_gen$
    \State $eva\_result = EVALUATE (eva\_pip)$
    \State $sel\_pip = SELECT (eva\_pip, eva\_result, N)$
    \State $iter\_init = sel\_pip$
\EndFor
\State Return $sel\_pip$
\Function{MUTATION}{$list\_pip$}
    \For {each integer $i$ in length of $list\_pip$}
        \State $list\_pip[i] = list\_pip[i]$  with one change in set (\ref{hyperparameterspace})
	\EndFor
\State Return $list\_pip$
\EndFunction
\Function{CROSSOVER}{$list\_pip$}
    \For {each pair ($pip\_1$, $pip\_2$) in $list\_pip$}
        \State Randomly separate $pip\_1$ into two parts.
        \State Randomly seprate $pip\_2$ into two parts.
        \State Combine the $1$st part of $pip\_1$ with the $2$nd part of $pip\_2$.
        \State Combine the $1$st part of $pip\_2$ with the $2$nd part of $pip\_1$.
        \State Update $pip\_1$ and $pip\_2$ in $list\_pip$
	\EndFor
\State Return $list\_pip$
\EndFunction
\Function{EVALUATE}{$list\_pip$}
    \State Train the $list\_pip$
    \For {each integer $i$ in length of $list\_pip$}
        \State $eva\_result[i] = CV(list\_pip[i])$
	\EndFor
\State Return $eva\_result$
\EndFunction
\Function{SELECT}{$eva\_pip, eva\_result, N$}
    \State $sel\_pip =$ the $N\ pips$ with highest $eva\_result$
\State Return $sel\_pip$
\EndFunction
\end{algorithmic}
\end{algorithm}

\subsection{Training and Testing Process}

This section presents the training and testing procedure.
The training process happens in the evaluation step as shown above.
Corresponding to our hierarchical framework, the pipeline is trained layer by layer.
Inside each layer, each primitive is also trained independently with the same dataset.
The next layer is trained on a concatenation of the previous dataset with the prediction results from the previous trained layer.
Similarly, the validation process and testing process share the same mechanism but with validation set and test set respectively. 

After training and validating the pipelines, we pick the first ten pipelines with the highest validation accuracies as the final output of Autostacker. We believe that these ten pipelines can provide better baselines for human experts to get started with the problem. Outputting a range of modeling options, rather than just the top single pipeline directly, allows the user more flexibility in her modeling process. We also consider the effect of small, unbalanced datasets; when taking in such datasets, it is difficult to guarantee that performance in the validation process can fully represent that on the test set. For example, two pipelines with the same validation results might behave very differently on the same test dataset. Hence, we consider it necessary to provide a set of candidates which can be guaranteed to do better on average so that human experts can fine tune the pipelines.

\subsection{Scaling and Parallelization}

Another significant advantage of our approach is that the system is very flexible to scale up and parallelize. This huge advantage comes for free when using Evolutionary Algorithms. Starting from the initial generation, one-step mutation, one-step cross-over, training, validation to evaluation, each pipeline runs independently, which means that each worker can work on one pipeline alone.

There is no frequent communication or sequential decision making among all the workers and each worker can run through the pipeline separately. Workers only need to share their validation results so they can be ranked by the end of each iteration. One shot selection, based on the validation accuracy, is subsequently applied on the outputs of the parallel workers. More specifically, in terms of the Algorithm \ref{phc} described above, Random(), 
MUTATION(), CROSSOVER(), and EVALUATE() function are all very easily parallelized when the system runs.

\begin{figure*}[hp]
\vspace{0mm}
\centering
    %\captionsetup[subfigure]{aboveskip=-8pt} % spaceing between image and caption for top row
	\begin{subfigure}[t]{1\linewidth}
		\centering
		\includegraphics[width=\linewidth,keepaspectratio]{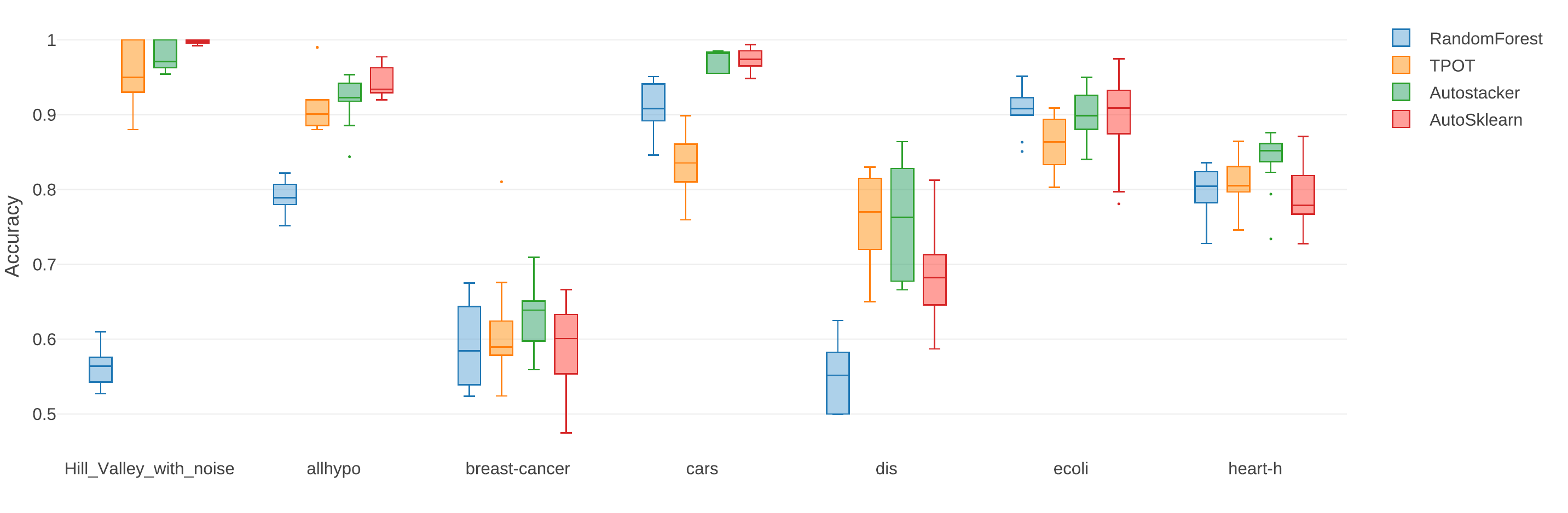}
		\label{fig:vision_1}
	\end{subfigure}
	\vspace{0.00mm}
	\begin{subfigure}[t]{1\linewidth}
		\centering
		\includegraphics[width=\linewidth,keepaspectratio]{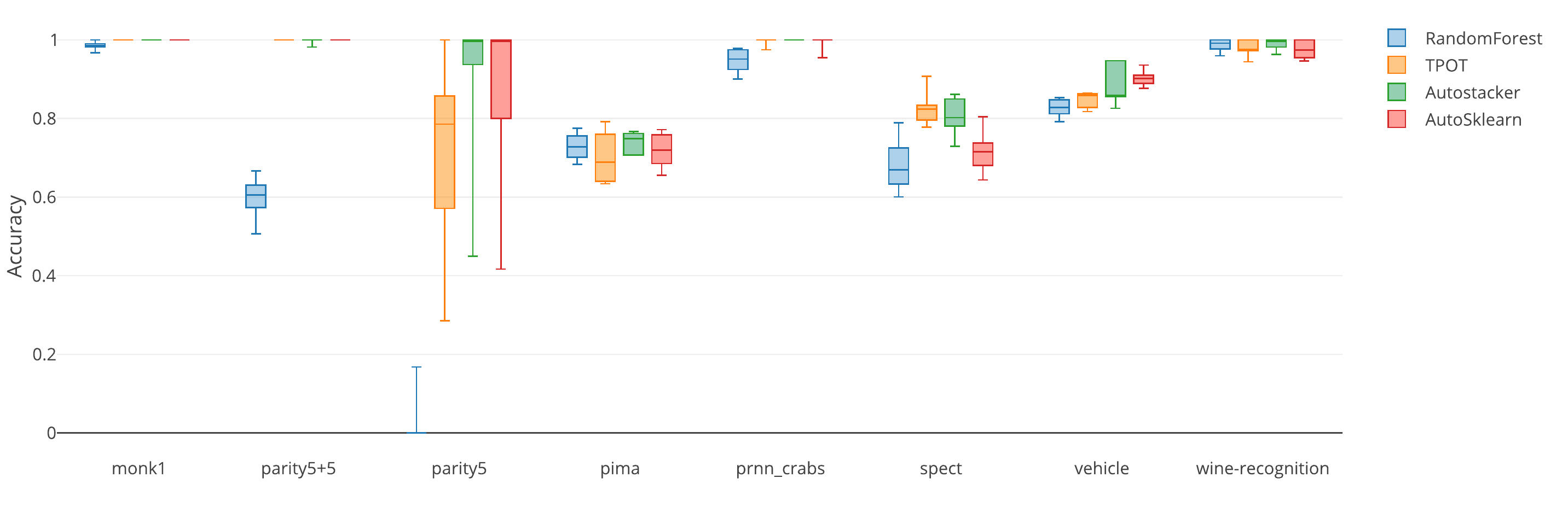}
		\label{fig:vision_2}
	\end{subfigure}
	\begin{subfigure}[t]{1\linewidth}
		\centering
		\includegraphics[width=\linewidth,keepaspectratio]{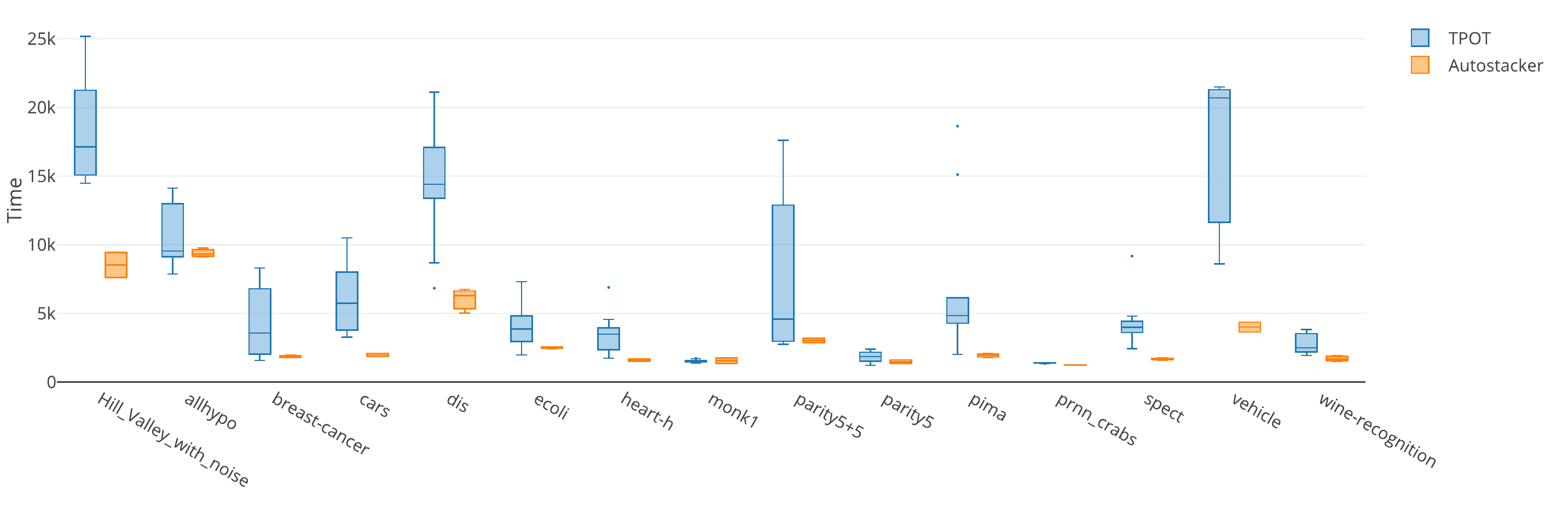}
		\label{fig:vision_3}
	\end{subfigure}

\caption{Test Accuracy and Time Cost Comparison.} 
\label{fig:results} 
\end{figure*}

\section{EXPERIMENTS}

\begin{figure}[h]
  \centering      
  \includegraphics[width=0.9\linewidth]{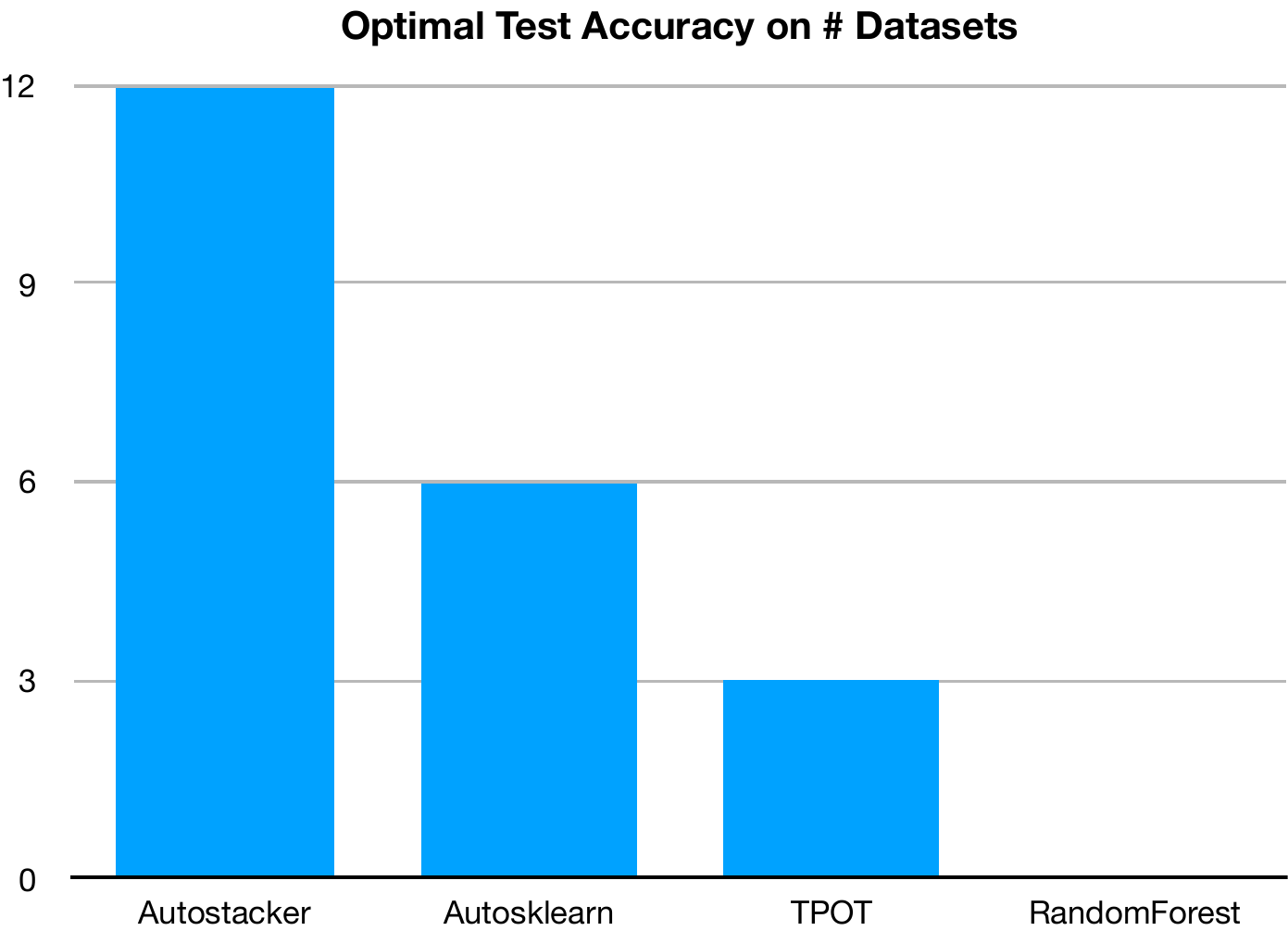}
  \caption{Autostacker outperforms all other architectures on average. The y-axis shows the number of datasets that each architecture outperforms all the other architectures.}
  \label{fig:bestperformance}
\end{figure}

\subsection{Dataset and Preprocessing}
To show the performance of our system, we select 15 datasets from the benchmark dataset provided in \cite{olson2017pmlb} which collects datasets from public data resources, such as OpenML \cite{OpenML2013} and UCI \cite{Lichman:2013} etc., as the sample experimental data. According to the result published in TPOT, we arbitrarily choose 9 datasets claimed to have better results in TPOT comparing with Random Forest Classifier, 4 datasets with worse performance in TPOT and 2 datasets with same performance with Random Forest Classifier in TPOT. We limit the total number of datasets to be 15 to show here to cover all cases of datasets used in TPOT. These datasets come from different problem domains and target different machine learning tasks including binary classification and multi-class classification.

There is no data preprocessing nor feature preprocessing currently involved in Autostacker. It would be certainly possible to use preprocessing on the dataset and features as another building block or hyperparameter in Autostacker, and we also provide this flexibility in our system. Nevertheless, in this paper we focus only on the modeling process to show our contribution to the architecture and automation process. Before each round of the experiment, we shuffle and partition the dataset to $80\%/20\%$ as training/testing data. Our code will be released.

\subsection{Baseline Comparison}

The goal of Autostacker is to automatically provide a better baseline pipeline for data scientists. Thus, the baseline we choose to compare with should be able to represent the prediction ability of pipelines coming from the initial trials of data scientists. The baseline pipeline that we compare with is chosen to be Random Forest Classifier with the number of estimators being 500 as ensemble learning models like Random Forest have been shown to work well on average in practice when considering multi-model predictions. We further compare our results to those of the TPOT model \cite{olson2016automating}, one of the more recent and popular AutoML systems, as well as AutoSklearn \cite{feurer2015efficient}, which won  1st place in the final phase of 2016 AutoML challenge \cite{guyon2016brief}. Both TPOT and AutoSklearn have open-sourced their systems. Hence, the current versions of these two systems have been improved a lot than the initial published versions by both the authors and the AutoML community. We use the most recent open-source versions of these two systems in our experiment.

\begin{table}[ht]
\centering
\caption{Primitive List in Autostacker}

\begin{tabular}{|l|l|}
\hline
Perceptron             & AdaBoostClassifier         \\ \hline
LogisticRegression     & XGBClassifier              \\ \hline
SVC                    & MLPClassifier              \\ \hline
DecisionTreeClassifier & BernoulliNB                \\ \hline
KNeighborsClassifier   & MultinomialNB              \\ \hline
RandomForestClassifier & GradientBoostingClassifier \\ \hline
BaggingClassifier      & ExtraTreesClassifier       \\ \hline
\end{tabular}
\label{primitive_list}
\end{table}

Currently, our primitives are from the scikit-learn library \cite{pedregosa2011scikit} and XGBoost library \cite{DBLP:journals/corr/ChenG16} as shown in Table \ref{primitive_list}. In Autostacker, users are allowed to plug in any primitives they like as long as the function signatures are consistent with our current code base. In terms of the basic structure (number of layers and number of nodes per layer) of the candidate pipelines, as we mentioned above, there are two types of settings provided in Autostacker. In this section, we show the performance of the default mode of Autostacker: dynamic configurations. We specify the maximum number of layers as 5 and the maximum number of nodes per layer as 3.

\subsection{Results}

In this section, we will show the results of the test accuracy and time cost of Autostacker as well as comparisons with the Random Forest, TPOT and AutoSklearn. The test accuracy is calculated using balanced accuracy \cite{velez2007balanced}. We refer to them as test accuracy in the rest of this paper. We ran 10 rounds of experiments for Random Forest Classifier and 3 to 10 rounds of experiment for TPOT based on the computation and time cost. For Autostacker, only 3 rounds of experiments are executed on each dataset and the datasets get shuffled before each round. The testing accuracies shown here come from the 10 top ranked pipelines outputted by Autostacker per round. Thus, the figure contains 30 test results in total. For AutoSklearn, we run 10 trials on each dataset with 1 hour time limitation for each round. The notches in the box plot represent the 95\% confidence intervals of median values. We ran our experiments using 24 CPU machines with 40GB of memory.

The first two rows in Figure \ref{fig:results} show test accuracy comparisons on the 15 sample datasets. We make several key observations based on the results:
\begin{itemize}
    \item Autostacker achieves \textbf{100\%} better test accuracy compared with Random Forest Baselines, \textbf{12 out of 15} better accuracy compared with TPOT, and \textbf{9 out of 15} better accuracy compared with AutoSklearn.
    \item Autostacker is robust - it provides a good baseline on every dataset. Random Forest fails to give meaningful results on the parity5 and parity5+5 datasets, and TPOT fails to provide better baselines than Random Forest Classifier on the breast-cancer, pima, ecoli, wine-recognition, and cars datasets after  multiple hours of computation time. AutoSklearn also fails to outperform any model on heart-h and wine-recognition.
\end{itemize}

The third row in Figure \ref{fig:results} shows the time cost of TPOT and Autostacker. We do not show the time cost of AutoSklearn here because it is mandatory to specify the time limitation of AutoSklearn beforehand or AutoSklearn will choose to use the default 1 hour as time limitation. We use this default setting in all of our experiments. As we show here, Autostacker largely reduces the time usage up to \textbf{6 times} compared to TPOT. Notably, TPOT seems to struggle with larger datasets (hill\_valley, dis, and parity5+5).  Autostacker also uses less time than AutoSklearn on 11 datasets. Interestingly, AutoSklearn outperforms Autostacker in both time and accuracy on three datasets (Hill\_Valley, allhypo, and vehicle). It is tempting to conclude that AutoSklearn performs better on larger datasets due to this observation, but we note that Autostacker had the highest test accuracy on the largest dataset (dis, 3772 samples). In terms of smaller datasets, however, Autostacker seems to have a natural advantage.

In conclusion of the experiment results summarized in Figure \ref{fig:bestperformance}, the output of Autostacker improves the baseline pipeline sufficiently enough for human experts to start with better pipelines within a short mount of time, and Autostacker outperforms all the baseline systems on average on all the sample datasets.

\section{DISCUSSION}

Despite great performance on the fifteen dataset benchmark, Autostacker still has several limitations. Here we will describe these limitations and possible future solutions:

\begin{itemize}

\item Many modern approaches and architectures achieve excellent results on large, high dimensional datasets and multi-task problems. Deep Learning, for example, has become a dominant approach in fields such as computer vision or natural language processing \cite{lecun}. Our current primitive library and modeling structure does not scale well to these problems. One direction of future work could be to incorporate more advanced primitives into Autostacker's catalog and to use them as necessary.

\item Autostacker can be made more efficient with better search algorithms. Many variants of evolutionary algorithms expand on the basic version used in our work. Experimenting with different algorithms may cause Autostacker to search faster or find better pipelines. We also believe a rigorous statistical analysis will help us better understand the output of Autostacker: why certain architectures are chosen, or how those architectures evolve over time.
\end{itemize}

\section{CONCLUSION}

In this work, we proposed Autostacker, an AutoML system inspired by stacking, cascading, and evolutionary algorithms. Despite the lack of data preprocessing and feature selection, Autostacker still outperforms competing AutoML systems on a wide variety of datasets in both accuracy and speed. We hope to provide a new benchmark in AutoML which bears the potential to incorporate more primitives and preprocessing techniques. 

\printbibliography
\end{document}